\documentclass{article}
\usepackage{spconf,amsmath,graphicx,hyperref}
\usepackage{booktabs,multirow}
\usepackage{amssymb}

\title{SKELETON-BASED SIGN LANGUAGE RECOGNITION USING A DUAL-STREAM SPATIO-TEMPORAL DYNAMIC GRAPH CONVOLUTIONAL NETWORK}
\name{Liangjin Liu, Haoyang Zheng, Zhengzhong Zhu,Pei Zhou$^*$\sthanks{$^*$Corresponding author: zhoupei@scu.edu.cn}}
\address{School of Computer Science\\
    Sichuan University\\
    Chengdu, China}

\begin{document}
\topmargin=0mm 
%
\maketitle

\begin{abstract}
Isolated Sign Language Recognition (ISLR) is challenged by gestures that are morphologically similar yet semantically distinct, a problem rooted in the complex interplay between hand shape and motion trajectory. Existing methods, often relying on a single reference frame, struggle to resolve this geometric ambiguity. This paper introduces Dual-SignLanguageNet (DSLNet), a dual-reference, dual-stream architecture that decouples and models gesture morphology and trajectory in separate, complementary coordinate systems. The architecture processes these streams through specialized networks: a topology-aware graph convolution models the view-invariant shape from a wrist-centric frame, while a Finsler geometry-based encoder captures the context-aware trajectory from a facial-centric frame. These features are then integrated via a geometry-driven optimal transport fusion mechanism. DSLNet sets a new state-of-the-art, achieving 93.70\%, 89.97\%, and 99.79\% accuracy on the challenging WLASL-100, WLASL-300, and LSA64 datasets, respectively, with significantly fewer parameters than competing models.
\end{abstract}

\section{Introduction}
\label{sec:intro}

Sign language recognition (SLR) is a critical technology for bridging the communication gap for hearing-impaired individuals \cite{Perumal_2024}. However, a fundamental challenge persists: accurately modeling the complex interplay of hand shape (morphology) and movement (trajectory). Many signs are morphologically similar but semantically distinct, where the meaning is defined by the hand's spatial relationship to the face or body. This ambiguity remains a significant hurdle for existing methods.
\begin{figure*}[!t]
    \centering
    \includegraphics[width=\textwidth]{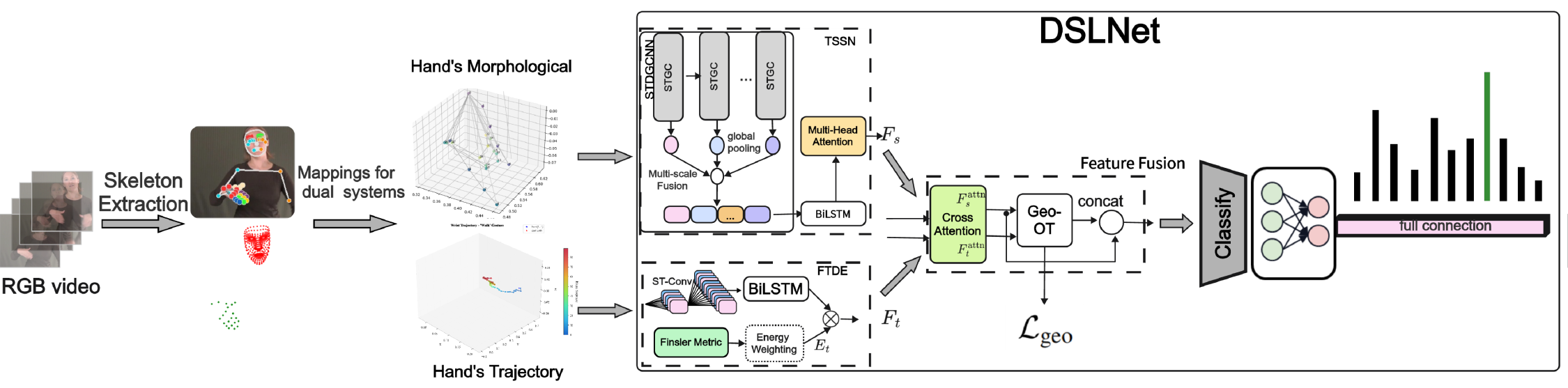}
    \caption{DSLNet Architecture. The left side shows the input skeletal data processed through dual reference frame mapping. The middle section features a dual-stream structure with shape and trajectory branches based on different reference frames. The right side includes a geometry-driven feature fusion module based on optimal transport theory and a classifier.}
    \label{fig:architecture}

\end{figure*}

Among classical gesture recognition methods, SAM-SLR \cite{jiang2021skeletonawaremultimodalsign} retains key points of the hands and upper body, introduces multi-stream inputs, and proposes SSTCN to process skeleton feature maps, thereby reducing the parameter count compared to traditional 3D convolutions. Based on this, SAM-SLR-v2 \cite{jiang2021signlanguagerecognitionskeletonaware} introduces a learnable late fusion module and incorporates 3D keypoints, which improves the handling of occlusion. The DSTA-SLR model \cite{hu2024dynamicspatialtemporalaggregationskeletonaware} breaks the fixed topology limitation of traditional GCNs by adopting an input-adaptive graph structure and introduces super nodes to enhance semantic understanding, while also employing parallel temporal convolution modules to capture multi-scale temporal dynamics. The ST-GCN-SL model \cite{101007} is the first to apply ST-GCN to sign language recognition. There are also models that combine ST-GCN with multi-cue LSTM for sign language recognition, such as MC-LSTM \cite{ozdemir2023multi}, which achieves adaptive temporal fusion of multiple cues.

Recent approaches have attempted to address these complexities with varying success. For instance, large-scale models like Uni-Sign \cite{li2025unisignunifiedsignlanguage} improve keypoint accuracy by pre-training on combined pose and visual data, but they struggle with temporal variations and lose spatial information when projecting 3D movements into 2D. Others, like NLA-SLR \cite{zuo2023naturallanguageassistedsignlanguage}, introduce linguistic context but remain sensitive to data imbalances and lack the fine-grained geometric understanding to differentiate spatially-dependent gestures. While computationally efficient models like SignBart \cite{nguyen2025signbartnewapproach} have emerged, they often sacrifice robustness by not fully integrating multimodal information.

To overcome these challenges, we propose DSLNet, a dual-reference, dual-stream architecture that decouples and models the two fundamental components of sign language: hand morphology and motion trajectory. 
DSLNet processes these inputs through two specialized streams: a topology-aware spatiotemporal network (TSSN) for morphological modeling and a Finsler geometry-based encoder (FTDE) for capturing direction-sensitive trajectory dynamics. These streams are then integrated using a geometry-driven optimal transport fusion mechanism that ensures semantic alignment between the shape and motion features.

Our contributions are threefold: (1) a novel dual-reference normalization strategy that provides complementary geometric constraints for robust recognition; (2) a specialized dual-stream architecture with components optimized for morphological and trajectory modeling; and (3) a geometry-driven fusion mechanism that achieves superior cross-modal alignment.

\section{Methodology}
\label{sec:methodology}
Given a skeleton sequence, our goal is to learn a mapping $f: \mathcal{X} \rightarrow \mathcal{Y}$ to a set of $C$ sign language classes. As illustrated in Figure~\ref{fig:architecture}, our DSLNet architecture achieves this through three key stages: dual-reference normalization, dual-stream feature extraction, and geometry-driven fusion.

\subsection{Dual-Reference Frame Normalization}
To resolve geometric ambiguities, we decouple the raw skeleton sequence into two semantically distinct representations. This is achieved by projecting the keypoint data onto two complementary, dynamically defined coordinate systems.

\noindent\textbf{Wrist Morphological Frame.}
To isolate the intrinsic hand configuration from global motion, we apply a translation operator $\mathcal{T}$ that re-centers the hand joint coordinates into a local, wrist-centric frame. This operator, parameterized by the wrist's position $h_w(t)$, generates a translation-invariant representation $\mathcal{X}'_{\text{shape}}$ that models pure morphology:
\begin{equation}
\mathcal{X}'_{\text{shape}}(t,i) = \mathcal{T}_{h_w(t)}(h_i(t)), \quad i \in \mathcal{J}
\end{equation}
where $\mathcal{T}_{\mathbf{p}}(\mathbf{x}) = \mathbf{x} - \mathbf{p}$ defines the translation.

\noindent\textbf{Facial Semantic Frame.}
To capture the gesture's contextual trajectory, we employ a normalization operator $\mathcal{N}$ that projects the wrist's global motion into a dynamic, body-centric coordinate system. This operator is anchored by a time-varying origin $c_f(t)$ and scale factor $s_f(t)$ derived from facial landmarks. The resulting representation, $\mathcal{X}'_{\text{traj}}$, preserves crucial spatial semantics while achieving pose and scale invariance:
\begin{equation}
\mathcal{X}'_{\text{traj}}(t) = \mathcal{N}_{c_f(t), s_f(t)}(h_w(t))
\end{equation}
where $\mathcal{N}_{\mathbf{c},s}(\mathbf{x}) = (\mathbf{x} - \mathbf{c}) / (s + \varepsilon)$ defines the normalization transformation.

\subsection{Dual-Stream Feature Extraction}
The two normalized inputs are processed by specialized streams designed to extract morphology and trajectory features, respectively.

\noindent\textbf{Morphology Stream (TSSN).}
The Topology-aware Spatiotemporal Network (TSSN), processes the shape representation $\mathcal{X}'_{\text{shape}}$. Its core is a Spatio-Temporal Dynamic Graph Convolutional Network (STDGCNN). This network consists of a stack of STGC blocks, where each block first performs spatial feature extraction on a dynamically constructed k-NN graph, followed by a temporal convolution to model local evolution. Features from multiple STGC blocks are aggregated to form a rich, multi-scale representation. The resulting feature sequence is further processed by a bidirectional LSTM and a multi-head attention layer to produce the final morphological feature $F_s$.

\noindent\textbf{Trajectory Stream (FTDE).}
The Finsler Trajectory Dynamics Encoder (FTDE) models the direction-sensitive dynamics of the trajectory $\mathcal{X}'_{\text{traj}}$. First, a learnable Finsler metric, which considers the trajectory's position, velocity, and direction, is used to compute a physics-informed temporal energy weight $E_t$ for each time step. Concurrently, the trajectory is processed by a Spatio-Temporal Causal Convolution (ST-Conv) network followed by a bidirectional LSTM to extract rich temporal features. The final trajectory representation $F_t \in \mathbb{R}^{T \times d_t}$ is obtained by modulating the output of the BiLSTM with the computed energy weights $E_t$, effectively emphasizing key moments in the gesture's execution.

\begin{equation}
F_t = \phi_\theta(p_t, \hat{v}_t) \cdot \|\dot{p}_t\|_2^{\alpha}, \quad E_t=\frac{F_t}{\sum_{\tau}F_\tau+\varepsilon}
\end{equation}
where $\phi_\theta$ is a fusion network and $\alpha$ is a learnable exponent. This energy $E_t$ acts as a physics-informed attention mechanism, emphasizing key moments like changes in direction or speed. The final trajectory representation $F_t \in \mathbb{R}^{T \times d_t}$ is obtained by modulating the output of a temporal convolution stack with these energy weights.

\subsection{Cross-Stream Fusion and Loss Function}
To integrate the two streams, we first enhance their features using a standard cross-attention mechanism. Then, a geometry-driven optimal transport (Geo-OT) module aligns the global morphological feature $F_s^{\text{attn}}$ with the temporal trajectory sequence $F_t^{\text{attn}}$. This is framed as finding an optimal transport plan $\gamma$ that minimizes a cost function combining feature similarity and temporal priors. The aligned trajectory feature $F_t^{\text{aligned}} = \sum_{j}\gamma_{1j} F_t^{\text{attn}}(j)$ is then concatenated with the morphological feature for classification.

The model is trained with a composite loss function:
\begin{equation}
\mathcal{L}=\mathcal{L}_{\text{CE}}+\alpha\,\mathcal{L}_{\text{geo}}
\end{equation}
where $\mathcal{L}_{\text{CE}}$ is the standard cross-entropy loss. The geometric consistency loss $\mathcal{L}_{\text{geo}}$ encourages semantic alignment between the two streams by maximizing the cosine similarity between their projected representations, ensuring that the learned features are consistent across modalities.
\begin{equation}
\mathcal{L}_{\text{geo}}=1-\cos\!\left(f_m(F_s^{\text{attn}}),\ f_a(F_t^{\text{aligned}})\right)
\end{equation}
Here, $f_m$ and $f_a$ are learnable projection heads.

\section{Experiments}
\label{sec:experiments}

We conduct a series of experiments to validate the effectiveness of DSLNet. 

\subsection{Implementation Details}
\textbf{Datasets.} We evaluate DSLNet on three widely-used datasets: LSA64\cite{lsa64} and WLASL\cite{wlasl}(WLASL-100 and WLASL-300).

\noindent\textbf{Preprocessing.} For each video, we extract 21 hand keypoints and 5 facial keypoints (nose and four mouth corners) accurately. The skeleton coordinates are normalized to the range $[-1, 1]$. We apply standard data augmentation, including random rotation, scaling, Gaussian noise, and temporal stretching to prevent overfitting.

\noindent\textbf{Training.} The model is implemented in PyTorch and trained on the NVIDIA RTX 4090 GPU. We use the AdamW optimizer with a cosine annealing learning rate schedule.

\subsection{Comparison with State-of-the-Art Methods}

\begin{table}[!h]
    \centering
    \caption{Performance comparison with state-of-the-art methods. Our model achieves the best accuracy on WLASL benchmarks with significantly fewer parameters.}
    \label{tab:sota_comparison}
    \footnotesize
    \setlength{\tabcolsep}{4pt}
    \begin{tabular}{llcc}
    \toprule
    \textbf{Dataset} & \textbf{Method} & \textbf{Params (M)} & \textbf{Acc (\%)} \\
    \midrule
    \multirow{5}{*}{LSA64} 
    & HWGAT\cite{jiang2021} & - & 98.59 \\
    & Siformer\cite{lin2023} & - & 99.84 \\
    & SPOTER\cite{bohacek2022} & 30.0 & 100.00 \\
    \cmidrule{2-4}
    & \textbf{DSLNet (Ours)} & 46.3 & 99.79 \\
    \midrule
    \multirow{5}{*}{WLASL-100} 
    & SignBERT\cite{hu2022} & - & 83.30 \\
    & NLA-SLR\cite{zuo2023} & 84.5 & 90.49 \\
    & Uni-Sign\cite{zhou2023} & 592.1 & 92.25 \\
    \cmidrule{2-4}
    & \textbf{DSLNet (Ours)} & 46.3 & 93.70 \\
    \midrule
    \multirow{5}{*}{WLASL-300} 
    & SignBERT\cite{hu2022} & - & 83.30 \\
    & NLA-SLR\cite{zuo2023} & 79.8 & 86.87 \\
    & Uni-Sign\cite{zhou2023} & 592.1 & 88.92 \\
    & SignBart\cite{nguyen2025signbartnewapproach} & 29.1 & 78.50 \\
    \cmidrule{2-4}
    & \textbf{DSLNet (Ours)} & 46.3 & 89.97 \\
    \bottomrule
    \end{tabular}
\end{table}

As shown in Table \ref{tab:sota_comparison}, DSLNet establishes new state-of-the-art performance on the challenging WLASL benchmarks and achieves competitive results on LSA64. On WLASL-100, DSLNet achieves \textbf{93.70\%} accuracy, outperforming the next best method, Uni-Sign, by a significant margin of 1.45\%. On the larger WLASL-300 dataset, our model reaches \textbf{89.97\%} accuracy, surpassing Uni-Sign by 1.05\%.

Crucially, DSLNet achieves this superior accuracy with remarkable efficiency. Compared to Uni-Sign, our model uses \textbf{12.8$\times$ fewer parameters} (46.3M vs. 592.1M). It is also more efficient than NLA-SLR, using 1.8$\times$ fewer parameters while delivering higher accuracy. This demonstrates that our architecture provides a superior trade-off between recognition performance and computational cost.

\subsection{Ablation Studies}
We conduct comprehensive ablation studies across all three datasets to validate our core design choices. The results consistently demonstrate the benefits of our proposed components.

\noindent\textbf{Effect of Dual-Reference Frames.} Table \ref{tab:reference_frames} validates our dual-reference strategy against a global normalization baseline. Our approach consistently yields superior performance, securing gains of \textbf{0.38\%} on WLASL-100 and \textbf{0.74\%} on WLASL-300. This confirms that decomposing gestures into view-invariant shape and context-aware trajectory provides a more potent feature space, effectively resolving geometric ambiguities for a more robust representation.
\begin{table}[!h]
\caption{Effect of Reference Frame Normalization. Accuracy (\%) is reported across all three datasets.}
\label{tab:reference_frames}
\centering
\footnotesize
\setlength{\tabcolsep}{3pt} 
\begin{tabular}{lccc}
\toprule
\textbf{Reference Frame} & \textbf{LSA64} & \textbf{WLASL-100} & \textbf{WLASL-300} \\ \midrule
Global Normalization  & 99.15 & 93.32 & 89.23 \\
\textbf{Dual Reference (Ours)} & 99.79 & 93.70 & 89.97 \\ \bottomrule
\end{tabular}
\end{table}

\noindent\textbf{Effect of Dual-Stream Architecture and Fusion.} Table \ref{tab:stream_analysis} analyzes the contribution of each stream and the fusion method. The morphology stream (TSSN) alone establishes a strong performance baseline (e.g., 87.45\% on WLASL-100), while the trajectory stream (FTDE) alone is insufficient. However, their combination is critical. Standard fusion methods like concatenation and cross-attention offer moderate gains. Our proposed geometry-driven optimal transport (Geo-OT) fusion, however, consistently achieves the best performance. It delivers a remarkable improvement of 2.16\% on WLASL-100 and 2.56\% on WLASL-300 over the standard cross-attention baseline. This highlights the superiority of Geo-OT in semantically aligning the two streams, especially for complex datasets.

\begin{table}[!h]
\caption{Analysis of Dual-Stream Components. Accuracy (\%) is reported across all three datasets.}
\label{tab:stream_analysis}
\centering
\footnotesize
\setlength{\tabcolsep}{3pt} 
\begin{tabular}{lccc}
\toprule
\textbf{Configuration} & \textbf{LSA64} & \textbf{WLASL-100} & \textbf{WLASL-300} \\ \midrule
TSSN only (Morphology) & 90.23 & 87.45 & 83.12 \\
FTDE only (Trajectory) & 65.61 & 43.33 & 50.98 \\
Dual-stream + Concatenation & 92.45 & 90.01 & 86.56 \\
Dual-stream + Cross-Attention & 94.12 & 91.54 & 87.41 \\
\textbf{Dual-stream + Geo-OT (Ours)} & 99.79 & 93.70 & 89.97 \\ \bottomrule
\end{tabular}
\end{table}

\subsection{Model Analysis}

\begin{figure}[h]
    \centering
    \includegraphics[width=0.5\linewidth]{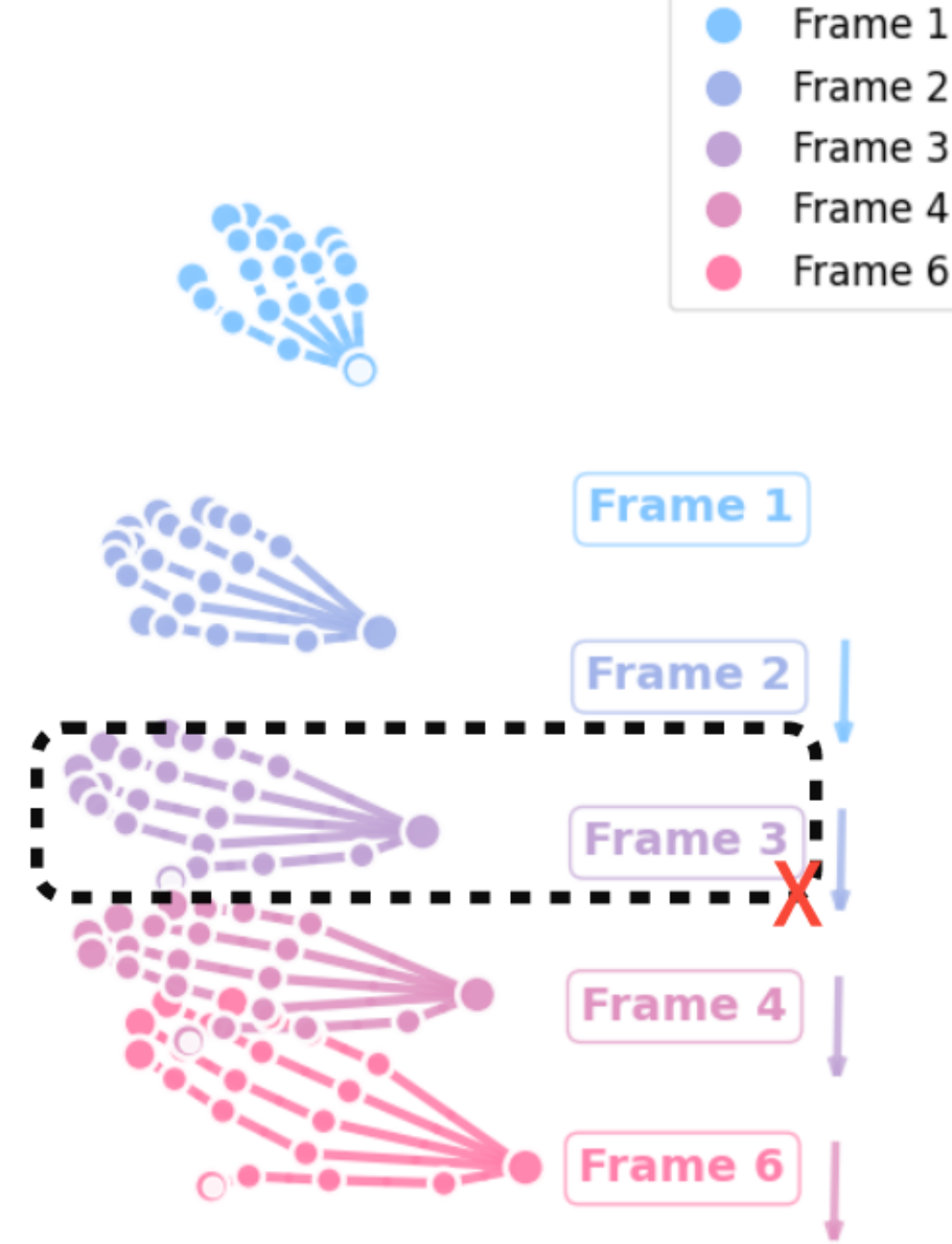}
    \caption{Frame dropout patterns used in robustness testing.}
    \label{fig:drop}
\end{figure}

\noindent\textbf{Robustness to Frame Dropout.} To simulate real-world conditions with data loss (e.g., motion blur, detection failure, Figure \ref{fig:drop})., we evaluate DSLNet's robustness by randomly dropping frames at varying rates. As shown in Table \ref{tab:robustness_analysis}, DSLNet demonstrates superior resilience compared to SPOTER. Even with a challenging 15\% frame dropout, DSLNet maintains a high accuracy of \textbf{85.23\%}, whereas SPOTER's performance degrades sharply to 76.45\%. This highlights the effectiveness of our FTDE trajectory stream, which can infer motion dynamics even from highly sparse temporal data.

\begin{table}[!h]
\caption{Robustness Analysis on LSA64 under Random Frame Dropout.}
\label{tab:robustness_analysis}
\centering
\footnotesize
\begin{tabular}{lcc}
\toprule
\textbf{Dropout Rate (\%)} & \textbf{DSLNet Acc. (\%)} & \textbf{SPOTER Acc. (\%)} \\ \midrule
0 (Baseline) & 99.79 & 100.00 \\
5 & 96.12 & 90.34 \\
10 & 93.56 & 85.12 \\
15 & 85.23 & 76.45 \\ \bottomrule
\end{tabular}
\end{table}

\noindent\textbf{Computational Efficiency.} DSLNet is designed for real-world deployment. It requires only \textbf{2.302G FLOPs} and achieves an average inference time of \textbf{17.98ms} per sample on an RTX 4090, well within real-time processing requirements (Less than 33ms). This efficiency, combined with its high accuracy, makes DSLNet a practical and powerful solution for ISLR.

\begin{figure}[h]
    \centering
    \includegraphics[width=0.7\linewidth]{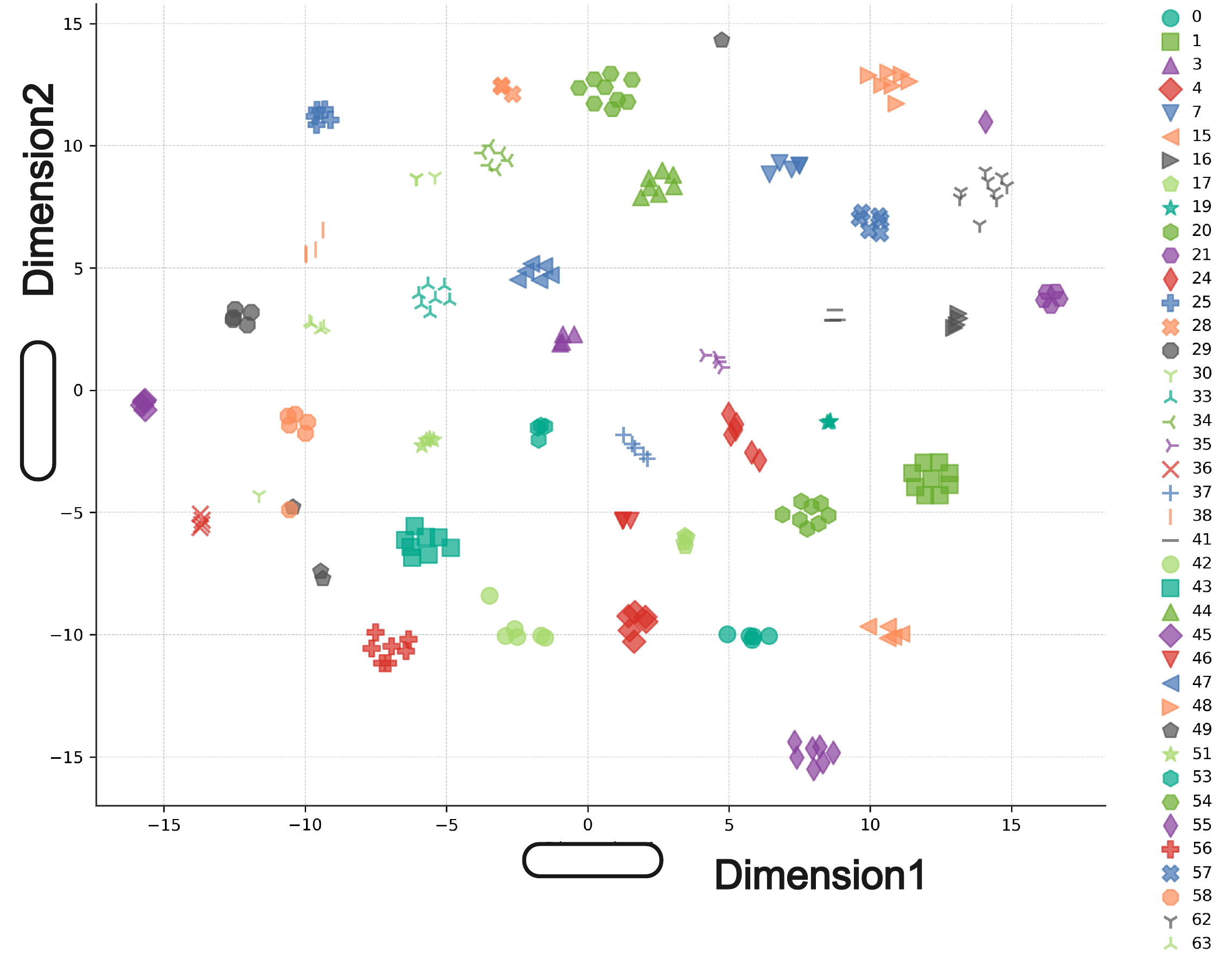}   
    \caption{t-SNE visualization of features learned by DSLNet on LSA64, showing distinct class clusters.}
    \label{fig:tsne}
\end{figure}

\noindent\textbf{Feature Visualization.} A t-SNE visualization of the learned features on LSA64 (Figure \ref{fig:tsne}) shows clear, well-separated clusters for different sign classes. This indicates that the model learns highly discriminative representations, which explains its strong classification performance.

\section{Conclusion}
\label{sec:conclusion}

This paper introduced DSLNet, a framework that resolves geometric ambiguities in skeleton-based sign language recognition by decoupling morphology and trajectory analysis. The synergy between a topology-aware shape network, a direction-sensitive trajectory encoder, and a geometry-driven optimal transport fusion mechanism leads to state-of-the-art performance. DSLNet achieves superior accuracy on challenging benchmarks with remarkable parameter and computational efficiency. This work validates the importance of multi-reference geometric modeling, offering a robust and practical solution for real-world ISLR. Future work will extend this framework to continuous sign language recognition and reduce dependency on precise keypoint detection.
\vfill\pagebreak

\bibliographystyle{IEEEbib}
\bibliography{refs}

\end{document}